%% file: main.tex
\pdfoutput=1

\documentclass[11pt]{article}

\usepackage{ACL2023}

\usepackage{times}
\usepackage{latexsym}

\usepackage[T1]{fontenc}

\usepackage[utf8]{inputenc}

\usepackage{microtype}

\usepackage{inconsolata}

\usepackage{amsmath}
\usepackage{array,graphicx}
\usepackage{color}
\usepackage{amsfonts}
\usepackage{latexsym}
\usepackage{url}
\usepackage{wrapfig}
\usepackage{comment}
\usepackage{booktabs}
\usepackage{microtype}
\usepackage{xspace}
\usepackage{booktabs}
\usepackage{multirow}
\usepackage{dcolumn}
\usepackage{hhline}
\usepackage{mdframed}
\usepackage{enumitem}
\usepackage{xcolor}
\usepackage{csquotes}
\usepackage{hyperref}
\usepackage{subfig}
\usepackage{caption}
\usepackage{microtype}
\newcolumntype{d}[1]{D{.}{.}{#1}}
\usepackage{makecell}
\usepackage{calrsfs}
\usepackage{ifthen}
\DeclareMathAlphabet{\pazocal}{OMS}{zplm}{m}{n}
\newboolean{showcomments}

\setboolean{showcomments}{false}  
\newcommand{\sof}[1]{\ifthenelse{\boolean{showcomments}}{{\color{red}sof:[#1]}}{}}
\newcommand{\jdcomment}[1]{\ifthenelse{\boolean{showcomments}}{{\color{blue}jesse:[#1]}}{}}
\newcommand{\nascomment}[1]{\ifthenelse{\boolean{showcomments}}{{\color{brown}nas:[#1]}}{}}

\title{Stubborn Lexical Bias in Data and Models}

\author{Sofia Serrano$^1$ \quad Jesse Dodge$^2$ \quad Noah A.~Smith$^{12}$ \\ $^1$Paul G.~Allen School of Computer Science \& Engineering, University of Washington \\ $^2$Allen Institute for Artificial Intelligence \\ \texttt{sofias6@cs.washington.edu, jessed@allenai.org, nasmith@cs.washington.edu}\\} 

\begin{document}
\maketitle
\begin{abstract}
In NLP, recent work has seen increased focus on spurious correlations between various features and labels in training data, and how these influence model behavior. However, the presence and effect of such correlations are typically examined feature by feature. We investigate the cumulative impact on a model of many such intersecting features.  Using a new statistical method, we examine whether such spurious patterns in data appear in models trained on the data. We select two tasks---natural language inference and duplicate-question detection---for which any unigram feature on its own should ideally be uninformative, which gives us a large pool of automatically extracted features with which to experiment. The large size of this pool allows us to investigate the intersection of features spuriously associated with (potentially different) labels. We then apply an optimization approach to \emph{reweight} the training data, reducing thousands of spurious correlations, and examine how doing so affects models trained on the reweighted data.   Surprisingly, though this method can successfully reduce lexical biases in the training data, we still find strong evidence of corresponding bias in the trained models, including worsened bias for slightly more complex features (bigrams).   We close with discussion about the implications of our results on what it means to ``debias'' training data, and how issues of data quality can affect model bias.
\end{abstract}

\section{Introduction}

Machine learning research today, including within NLP, is dominated by large datasets and expressive models that are able to take advantage of them.  At the same time, as the scale of training data has grown, this explosion of data has come at the expense of data \emph{curation}; for many of the datasets currently in use today, human oversight of the full breadth of their contents has become unrealistic. This makes it more likely that training datasets contain undesirable associations or shortcuts to learning intended tasks.  Many cases are attested \citep[e.g.,][]{Tsuchiya2018PerformanceIC, gururangan-etal-2018-annotation, Poliak2018CollectingDN, mccoy-etal-2019-right, rudinger-etal-2018-gender, stanovsky-etal-2019-evaluating, davidson-etal-2019-racial, sap-etal-2019-risk}, and we suspect a vast number of these so-called ``spurious correlations'' remain undetected.  

One question is whether these unintended biases in the training data propagate to models trained on that data. Recent work has found mixed results on this point \citep{steed-etal-2022-upstream, joshi-he-2022-investigation}. We begin by introducing an approach to testing for undesirable model biases that can operate using existing held-out data, even though that data might itself have spurious correlations.  In particular, we repurpose the classic permutation test to examine whether observed differences in model performance between instances exhibiting more common feature-label pairings and those exhibiting less common feature-label pairings are statistically significant.

For our experiments, we focus on the simplest kind of feature-label association:  correlations between lexical features and task labels. We select two tasks (natural language inference and duplicate-question detection) for which any such lexical feature should be uninformative on its own.   Finding strong evidence that models finetuned on three different datasets have at least some of the same lexical biases that exist in their training data, we then examine the extent to which those biases are mitigated by lessening biases in the training data. To do this, we apply an optimization-based approach to reweighting the training instances.  The approach brings uneven label distributions closer to uniform for thousands of different intersecting lexical features, many more than we use for our model bias evaluation, and still manages to have a strong effect on the most initially biased features despite our reweighting approach not focusing on those in particular.  We then finetune new models on those (reweighted) datasets. We find that although model bias lessens somewhat when we do this, we still find strong evidence of bias.   Surprisingly, this holds even when we consider models that make use of no pretraining data.

We close with a discussion of possible factors contributing to these results. We first note that perhaps the continued relative lack of variety of minority-class examples containing certain features hinders the reweighted models' ability to generalize their recognition of those less-common feature-class pairs, even though the combined weight given to those few instances in the loss function is increased. However, when we examine the effect of our reweighting on higher-order features (namely, bigrams), we see another problem: the same reweighting that mitigates associations between unigrams and any particular label actually strengthens associations between bigrams and certain labels in data. Based on this observation, we arrive at two conclusions: (1) simultaneously reducing bias across features of different levels of granularity for natural-language data is likely not feasible, and (2) even if we aim to mitigate model bias \textit{only} with respect to simple features, if we do so by reweighting the data, the high-capacity models used in modern NLP are still capable of learning the spurious correlations of the original unweighted data through associations that remain encoded in more complex features even after reweighting. We conclude that bias reduction in NLP cannot be cast purely as a ``data problem,'' and solutions may need to focus elsewhere (e.g., on models).

\section{What Do We Mean by Bias?} \label{sec:biasdef}
The term ``bias'' is polysemous, having been adopted by different communities to mean different things, from historically rooted social inequity to skewed model evaluations \cite{mehrabi2021survey} to techniques that help with supervised class imbalance in labels \cite{chen-etal-2018-collective}.
In our work, we use ``bias'' to mean correlations between individual input features and task labels. 
This framework is fairly general, but our focus in this work is natural language data. Therefore, as an example to illustrate our definition of bias, we will refer to correlations between the presence of individual word types in the input (unigrams) and a given label in a classification task.

More formally, consider a task of mapping inputs in $\pazocal{X}$ to labels in $\pazocal{Y}$.  We assume a training dataset $\pazocal{D} = \langle (x_i, y_i)\rangle_{i=1}^n$, each $x_i\in\pazocal{X}$ and $y_i\in\pazocal{Y}$. We are particularly interested in a designated collection of $d$ binary features on $\pazocal{X}$, the $j$th of which is denoted $f_j : \pazocal{X} \rightarrow \{0,1\}$. For example, $f_j$ might be the presence of the word ``nobody'' in an instance. Let $f_{j,i}$ be shorthand for $f_j(x_i)$ (e.g., whether instance $x_i$ contains the word ``nobody'' ($f_j(x_i) = 1$) or not ($f_j(x_i) = 0$)).

Introducing random variable notation, we can characterize $\pazocal{D}$ by its empirical conditional distribution over labels given each feature, such that for all $y \in \pazocal{Y}$,
\begin{align}
    \hat{p}(Y = y \mid F_j = 1) 
 &= 
\frac{\sum_i \boldsymbol{1}\{f_{j,i} = 1 \wedge y_i = y\}}{\sum_i \boldsymbol{1}\{ f_{j,i} = 1 \}}. \nonumber
\end{align}
If the conditional distribution of output labels given the presence of a particular lexical feature is very different from the overall label distribution in the data, we consider that feature to be biased in the training data.

\section{Measuring Bias in Model Performance and Data}
Recall that when $\hat{p}(Y = y \mid F_j = 1)$
is close to $1$, it means feature $j$ is correlated with label $y$ in a given dataset.
Let us denote the set of examples that contain feature $j$ and have 
the label most strongly associated with feature $j$ in $\pazocal{D}$ by $\pazocal{U}_{j}$, which we call the ``usual-labels'' set.
Then, denote the examples that contain $j$ but have a \emph{different} label by $\pazocal{N}_{j}$, which we call the ``unusual-labels'' set.

To build intuition, the accuracy of the model on instances which contain feature $j$ is the accuracy over the union $\pazocal{U}_j \cup {\pazocal{N}_{j}}$.
However, to measure if the model is picking up bias from the data, we will measure accuracy over $\pazocal{U}_{j}$ and $\pazocal{N}_{j}$ separately.
To maximize accuracy on $\pazocal{U}_{j} \cup {\pazocal{N}_{j}}$ the model would be justified in disproportionately labeling instances containing $f_j$ with $y$, so we can't  use accuracy by itself to measure model bias.
Instead, the key idea here will be to look for differences in error rates between instances whose labels align with features' training biases (the ``usual-labels'' set), and instances whose labels do not.

If the model has learned a biased representation of the data, we expect it to have higher accuracy on the ``usual-labels'' set, $\pazocal{U}_{j}$.
On the other hand, if the model hasn't learned that bias, we would expect the correct predictions to be uniformly distributed between $\pazocal{U}_{j}$ and $\pazocal{N}_{j}$.
We use this as the basis for a hypothesis test:
the null hypothesis $H_0$ is that the accuracy of model is the same on both sets $\text{ACC}(\pazocal{U}_{j})=\text{ACC}(\pazocal{N}_{j})$, and the alternative hypothesis $H_1$ is that $\text{ACC}(\pazocal{U}_{j})>\text{ACC}(\pazocal{N}_{j})$.
That is, if the errors are distributed uniformly at random, how likely is it that $\pazocal{U}_{j}$ would have \textit{at least} its observed number of correct instances?

\subsection{Permutation Test}

Given a model's accuracy on $\pazocal{U}_j$ and $\pazocal{N}_j$, and the size of the two sets, we can calculate the $p$-value for this hypothesis test exactly using the permutation test \citep{PhipsonSmyth+2010}. Our null hypothesis is that the errors are uniformly distributed between $\pazocal{U}_j$ and $\pazocal{N}_j$, so the permutation test calls for randomly shuffling whether a given instance is correctly labeled, while not changing the number of instances in each category \emph{or} the model's overall accuracy on the set union, both of which change the shape of the distribution of correct instances that we'd expect to see, but neither of which is the property for which we're testing. As there are finitely many ways to shuffle whether a given instance is correctly labeled, this test also has the benefit of having a closed form, giving us an exact $p$-value.\footnote{For simplicity, we assume here that the model has an equal likelihood of guessing any of the output classes. In practice, this is approximately accurate for the data on which we experiment, though this assumption could be removed in principle by multiplying each permutation by a corresponding probability.
}

\subsection{Calculating Bias over Multiple Features}
\begin{figure*}[!ht]
  \centering
  \includegraphics[width=\textwidth]{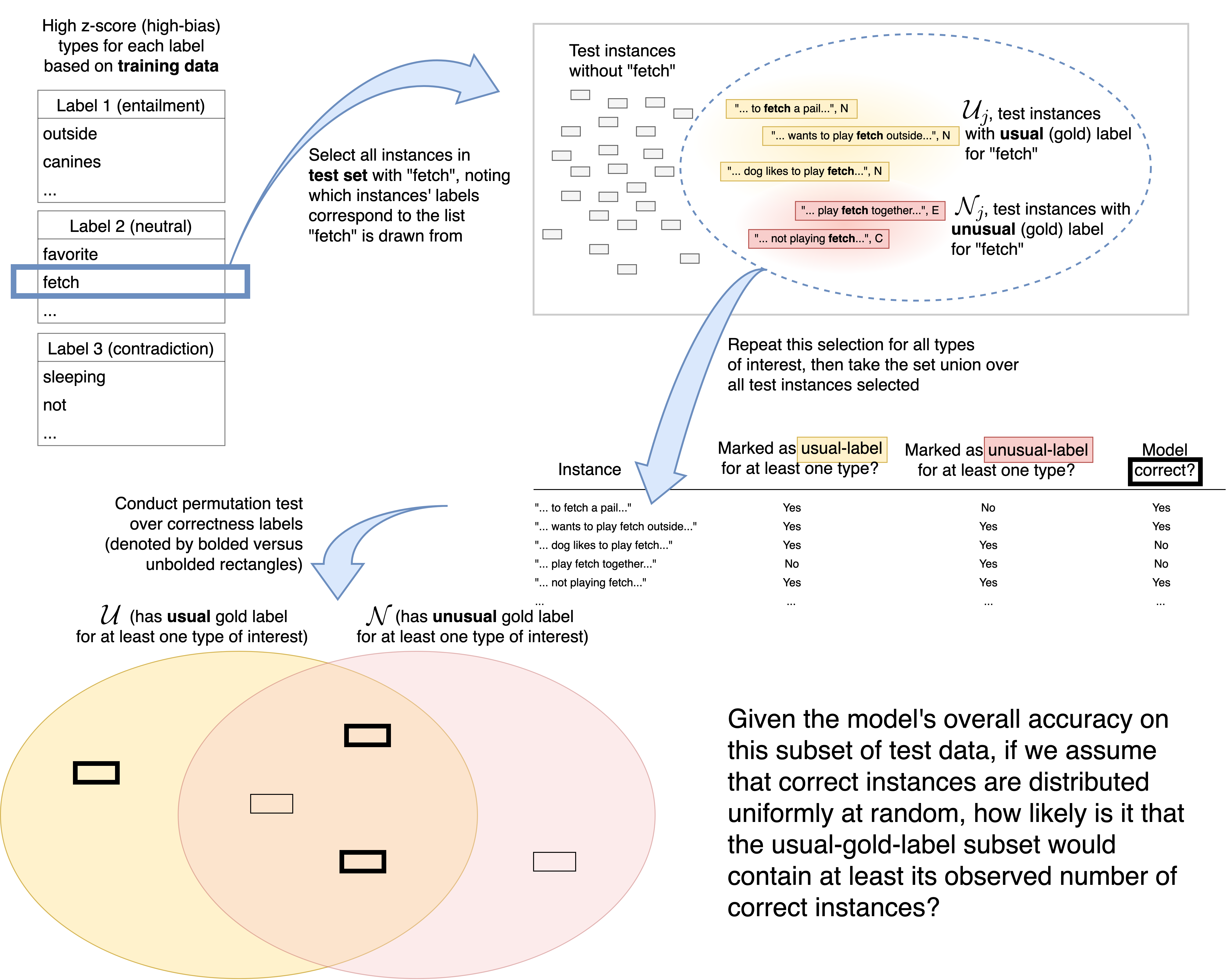}
  \caption{The setup of the permutation test that we use to test for bias in models trained on biased data, which in this figure uses word types as features and natural language inference as the underlying task.}
  \label{fig:permtestsetup}
\end{figure*}

In the previous section we described how we could use a permutation test for a single feature $f_j$.
Here we describe how to apply this to the full dataset.
We define $\pazocal{U}$ as $\cup_j \pazocal{U}_j$ and $\pazocal{N}$ as $\cup_j\pazocal{N}_j$ for 50 features $f_j$ per distinct label (namely, those that demonstrate the highest association with that label in the training data), so 100 or roughly 150 features $f_j$ total depending on whether the dataset is 2- or 3-class (``roughly'' because some features are among the most associated for two classes in 3-way classification). 
Given that each example $x_i$ includes multiple features (e.g., $f_{j,i}=1\wedge f_{k,i}=1$) it's possible for example $x_i$ to have label $y$, which is the ``usual-labels'' for $f_j$ but an ``unusual-labels'' for $f_k$.
When this happens, we add it to both sets $\pazocal{U}$ and $\pazocal{N}$, meaning that their intersection is not necessarily empty.
Pooling examples in this way allows us to run a single hypothesis test for whether or not the model learns bias from the dataset, avoiding the multiple-comparisons issue of running one hypothesis test for each feature.
This procedure is described in Figure~\ref{fig:permtestsetup}.

\section{Applying the Test}

Here we shift our focus to particular tasks and datasets, in order to apply our test in practice.

\subsection{Determining Biased Features (and Tasks)}

For our experiments, we want a large volume of features that should ideally exhibit no correlation with labels. In order to get a large number of features, we'd like them to be simple and easy to automatically detect, so unigram features again come to mind, guiding our selection of tasks and datasets for experiments.

When is the association of unigram features with a particular label a problem? While previous work has argued that the presence of an individual word type in a given instance, by itself, does not provide enough information to predict the label for \textit{any} ideal task that requires an understanding of natural language \cite{gardner-etal-2021-competency}, in this work we consider this argument only as it relates to two tasks where such a position is relatively uncontroversial: natural language inference, and duplicate-question detection.

Consider the task of natural language inference (NLI), where the input consists of two sentences (premise and hypothesis), and the correct label is a human annotation indicating whether the premise entails the hypothesis, contradicts it, or neither.
Continuing our example from section ~\ref{sec:biasdef}, if $f_{j,i}=1$, then the word ``nobody'' appears somewhere in example $x_i$ (premise, hypothesis, or both).
Given these definitions of the task and the features, $f_{j,i}=1$ by itself is uninformative for predicting $y_i$ (intuitively, we don't learn any information about whether or not the premise entails the hypothesis by knowing that the word ``nobody'' appears somewhere in the input).
However, it has been shown that in the SNLI dataset \cite{bowman-etal-2015-large} $f_{j}=1$ almost perfectly predicts the label, in both the training and test sets (for example, in the training set, 2368 instances with $f_j = 1$ have a label of ``contradiction'' and only 13 don't). 
Thus, this is an example of a ``spurious correlation'' (or, bias in the data).

\subsection{Applying the Test to Models}

We now apply the described permutation test to finetuned models. For each of SNLI \citep{bowman-etal-2015-large}, QNLI \citep{wang-etal-2018-glue}, and QQP,\footnote{Quora Question Pairs dataset (QQP): \url{data.quora.com/First-Quora-Dataset-Release-Question-Pairs}} we finetune three pretrained RoBERTa-large models \citep{roberta} with different random seeds on their training sets. We use a learning rate of $2\times 10^{-6}$ and finetune for 15 epochs using a single GPU with 12GB memory.

Following the argument by \citet{gardner-etal-2021-competency} that unigram features for these kinds of theoretically complex tasks should ideally be uninformative in isolation, we use lexical types as our bias evaluation features. 
For the purpose of this calculation, each label will contribute the 50 features that have the strongest correlation with it (as calculated by $z$-score, again following \citealp{gardner-etal-2021-competency}) in the lowercased training data, excluding stop words, since they tend to receive high $z$-scores due to appearing in such an overwhelming number of instances.\footnote{In section~\ref{sec:snli-words}, for illustration purposes, we include the resulting list of 50 lexical types per label for SNLI.} We then select all test instances with one or more of those types present as our evaluation set for our permutation test. For models finetuned on SNLI and QQP, we find $p$-values of at most $2.3 \times 10^{-17}$ (see ``Trained on uniform'' rows of Table~\ref{tab:pvals}), indicating very strong evidence that---as expected---these models reflect the bias associated with types with high $z$-scores in the training set. For QNLI, we see mixed results depending on our random seed, with $p$-values of 0.0057, 0.024, and 0.053 for our three finetuned models. (Worth noting is the fact that, as we will see later in Section~\ref{sec:intervening-balancing}, QNLI has the lowest overall feature-label bias of any of these three datasets.) Still, we see enough of these models demonstrating bias to merit investigating why this occurs.

\section{Where Does that Bias Come From?}

Having established that there is often similar bias in the finetuning data and models trained on that data, we consider that the finetuning data is not necessarily the source of the bias in the model. For example, the bias could come from the pretraining data as well. With that in mind, how might we check the impact of the finetuning data specifically?

\subsection{Intervening on the Data by Balancing It} \label{sec:intervening-balancing}

Our strategy is to intervene on the data to lessen lexical bias.\footnote{Note, we do not describe our approach as ``removing bias,'' as natural language data in general is biased to some extent; see the argument made by \citet{schwartz-stanovsky-2022-limitations}.} While modifying the data is only one family of approaches towards reducing eventual bias of a learned model (see, for example model-based strategies such as those proposed by \citealp{clark-etal-2019-dont}, or \citealp{karimi-mahabadi-etal-2020-end}), recall that our goal here is to investigate the effect of the finetuning data on the rest of the training setup, so for our purposes we keep the rest of the training procedure the same.

Prior work has explored different ways of intervening on data, such as manual data augmentation \citep{zhao-etal-2018-gender, Zhang2020TowardsAP, Gowda2021PullingUB, Lee2021CrossAugAC}, or occluding bias in the original data \citep{feldman-certifyingremoving}, but along very few different axes of bias. Other work augments minority-class data for the purpose of addressing class imbalance \citep{chawla-smote}. Yet others have taken the approach of generating new data to augment the existing data in ways that counteract certain biases \citep{wu-etal-2022-generating}. However, this last work relies on model-generated text, which, as \citet{wu-etal-2022-generating} themselves acknowledge, could differ from human-generated text in ways that aren't immediately obvious \citep{zellers-grover}. 

In order to avoid potential new artifacts introduced by using machine-generated training data, and to improve the label balance in aggregate for a large volume of features simultaneously, we reweight existing training data such that in expectation, the disproportionate association of lexical features with certain labels is decreased. Reweighting data to remove bias is not a new idea---\citet{kamiran2012data} do this through downsampling---but typically such approaches have considered at most a handful of different axes of bias. Some existing work, namely \citet{Byrd2018WhatIT} and \citet{Zhai2022UnderstandingWG}, has pointed out the limitations of approaches based on reweighting data, but again based on reweighting along comparatively few axes (in the case of the former) or on simpler model architectures than we consider here (in the case of the latter), so in the absence of a viable alternative meeting our requirements, we proceed with reweighting as our form of intervention for our experiments.

Typically, training datasets like $\pazocal{D}$ are treated as i.i.d., representative samples from a larger population.  Formally, we instead propose to \emph{weight} the instances in $\pazocal{D}$, assigning probability $q_i$ to instance $i$, such that, $\forall j, \forall y \in \pazocal{Y},$
\begin{align}
\frac{\sum_i q_i \cdot \boldsymbol{1}\{f_{j,i} = 1 \wedge y_i = y\}}{\sum_i q_i \cdot \boldsymbol{1}\{ f_{j,i} = 1 \}} & = \frac{1}{|\pazocal{Y}|} \label{eq:goal}
\end{align}
From here on, we denote the lefthand side of Equation~\ref{eq:goal} as $q(y \mid F_j=1)$.  Note that, for simplicity, we assume a uniform distribution over labels as the target, though our methods can be straightforwardly adapted to alternative targets.

Given an algorithm that produces a weighting $q_1,\ldots,q_n$ for dataset $\pazocal{D}$, we quantify its absolute error with respect to Equation~\ref{eq:goal} as
\begin{align}
    \textrm{Err}(q) \ =& \ \frac{1}{\textrm{(number of features}) \cdot |\pazocal{Y}|}  \ \cdot  \nonumber \\
    &  \ \sum_j \sum_{y \in \pazocal{Y}} \left| q(y \mid F_j=1) - \frac{1}{|\pazocal{Y}|}\right| \nonumber
\end{align}

How do we choose these $q_i$ values? We can state the general problem as a constrained optimization problem.\footnote{The slightly simplified formulation we present here for ease of reading only takes into account cases where feature $j$ appears somewhere in our data, but Equation~\ref{eq:balance} can be straightforwardly modified by multiplying it by the denominator of $q(y \mid F_j = 1)$ to account for this.}
We seek values $q_1, \ldots, q_n$ such that:
\begin{align}
    \sum_{i=1}^n q_i &= 1 \label{eq:sum-to-one} \\
    q_i & \ge 0, \ \forall i  \label{eq:nonneg}\\
q(y \mid F_j = 1) - \frac{1}{|\pazocal{Y}|} &= 0, \ \forall j, \forall y \in \pazocal{Y} \label{eq:balance}
\end{align}
(The constraints in the last line are derived from Equation~\ref{eq:goal}; strictly speaking one label's constraints are redundant and could be removed given the sum-to-one constraints.)

Using this setup, we seek a vector $q$ that satisfies the constraints.  We do this by minimizing the sum of squares of the left side of Equation~\ref{eq:balance}; the approach is simplified by a reparameterization:
\begin{align}
    q_i &= \frac{\exp z_i}{\sum_i \exp{z_i}} \nonumber
\end{align}
This is equivalent to optimizing with respect to unnormalized weights ($z_i$) that are passed through a ``softmax'' operator, eliminating the need for the constraints in Equations~\ref{eq:sum-to-one} and~\ref{eq:nonneg}. Once we have $q$, we multiply each $x_i$'s contribution to the loss during training by $q_i \cdot |\pazocal{D}|$.

\input{figs/table-balancingdatasets}

We apply this algorithm to reweight the following training datasets: SNLI \cite{bowman-etal-2015-large}, MNLI \cite{williams-etal-2018-broad}, QNLI \cite{wang-etal-2018-glue}, and QQP. In contrast to the <200 features per dataset that we use for evaluation of bias in models, when reweighting data, we used all types that appeared at least 100 times in their corresponding training data as features, and we denoted an ``instance'' as the concatenation of a paired premise and hypothesis (or, for QQP, the concatenation of the two questions). We removed features from consideration if they did not have at least one document in the dataset for each of their labels.\footnote{This was not the case for any features in MNLI or QNLI, but applied to the word ``recess'' for SNLI, and the words ``gobi'' and ``weakest'' for QQP.}

\begin{figure}[!ht]
  \centering
  \includegraphics[width=0.5\textwidth]{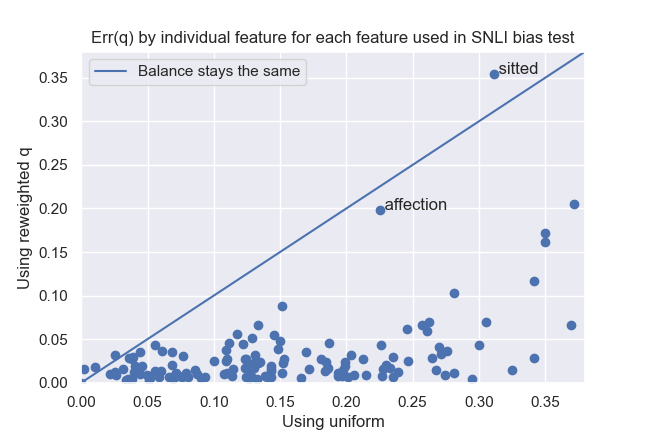}
  \caption{Label balance of the 137 lexical features used in our \textit{model} bias evaluation for SNLI (since a handful of the highest $z$-score features in the training data didn't appear in the test set), using a uniform weighting and reweighed using $q$. $q$ produces a lower Err($q$) for most of these features and is comparable for most of the remaining few, even considering that the reweighting was with respect to all 3,866 features. We have labeled the only two features that go against this pattern.}
  \label{fig:featbalance}
\end{figure}

We see in Table~\ref{tab:rebalancing-real-data} that by solving for distributions $q$ over the different datasets as described, we successfully reduce $\textrm{Err}(q)$ compared to the initial uniform weighting for all datasets except MNLI.\footnote{MNLI is unusual among the datasets we studied in its remarkably low degree of lexical-feature bias to begin with, so it is perhaps not surprising that further lowering that bias across thousands of features proves difficult.} This leaves us with three successfully reweighted datasets with lessened unigram bias overall, and we can use these to investigate possible reduction of lexical bias compared to their original, uniformly-weighted counterparts. We confirm that for the high-$z$-score features used for model bias evaluation for each of these three, their label balance in the data either improves (often dramatically) or stays comparable as a result of our reweighting $q$. (Here and elsewhere, we use ``label balance'' of a feature to refer to the average absolute difference between its empirical label distribution in the training data and the overall label distribution of the training data, averaging elementwise over each possible label.) For example, see Figure~\ref{fig:featbalance} for the change that our reweighted $q$ makes in improving the label distributions of our original high-$z$-score features from SNLI that we use for evaluation.

\subsection{Impact when Finetuning on Reweighted Data}

We now consider what happens when we finetune models on that data. We finetune RoBERTa-large models using new random seeds and all the same hyperparameters as before, only this time on training data reweighted using the new $q$ distributions. We see similar validation accuracies (a point or so of difference), indicating that this reweighting has a small effect on overall performance, even though the validation sets may contain similar biases to their corresponding training sets and therefore benefit models that leverage those biases.

\input{figs/table-pvals}

The results of rerunning our model bias evaluation are listed in the top half of Table~\ref{tab:pvals}. While we do see an increase in $p$-values, indicating weaker evidence of bias than for models trained on the uniformly-weighted training data, for both SNLI and QQP, we are still left with very strong evidence of bias ($p$-values of at most $1.2\times 10^{-5}$). A natural question that we might ask is whether we can attribute this remaining bias to the pretraining data.  

To test whether we see the same patterns in the absence of any other training data, we also train two bidirectional three-layer LSTMs per dataset from scratch (i.e., no pretraining and no pretraining data), one using uniform weighting and the other using $q$-reweighted.\footnote{To ensure no leaked signal from any other data, we initialized the word embeddings of the LSTMs to continuous bag-of-words embeddings \citep{mikolovword2vec} trained using their respective $q$-weighted training sets. We use a word embedding dimension of 128, a hidden size as input to the second LSTM layer of 256, and a hidden size as input to the third LSTM layer of 512. That third layer outputs a 128-dimensional vector, to which a linear projection projecting it to the appropriate number of output dimensions is then applied.} As we can see in Table~\ref{tab:pvals}, while there continues to be a rise in $p$-value with the switch to the reweighted $q$, the higher $p$-value is still vanishingly small.  \textbf{All the models trained from scratch are biased.}

Of particular interest is the fact that the LSTMs trained on QNLI display strong evidence of bias, while the pretrained transformers that were finetuned on either version of QNLI (reweighted or not) were the only models that did not display strong evidence of bias. This indicates that at least in QNLI's case, bias has entirely separate causes than training data; for QNLI, it's only the models trained from scratch that display significant evidence of bias. This, along with the tiny $p$-values for the other LSTMs, indicates that there are still factors even in the reweighted data that contribute to bias.

\input{figs/table-bigrambalance.tex}

At first, this is surprising. Given that the LSTMs trained with the reweighted $q$ distributions over data were exposed to no other data, why do they still exhibit bias? One possibility is issues of quality inherent to some unusual-label data. For example, consider the word ``favorite'' in SNLI, which has one of the highest $z$-scores for the ``neutral'' label. Even though nothing about the task of determining whether one sentence entails another inherently suggests an association between ``favorite'' and a particular label, since SNLI was constructed based on photographs (without any additional data about their subjects' mental states) as the underlying source of data for written premises, we expect the term ``favorite'' to occur mostly in hypotheses that are neither entailed nor contradicted by this data. Even though the reweighted $q$ gives more weight to unusual examples, those examples could sometimes be of lower quality due to details of how the data was collected. 

Furthermore, even though the total contribution to the loss function during training is approximately the same across labels using the reweighted $q$, the model still sees a wider variety of instances for types' ``usual'' labels, which perhaps allows it to generalize better in that regard. In other words, the characteristics of less common ($f_j, y$) pairings aren't inherently easier for a model to learn than the characteristics of more common pairings, so models' generalization to new examples with the less common $(f_j, y)$ pairing would still be hurt by seeing a smaller variety of examples representing those kinds of instances, even if that smaller variety received greater total weight in the loss function.

\section{Effects of  Rebalancing on Higher-Order Features}

We have found that rebalancing labeled data doesn't remove bias in a downstream model.  Another possible explanation is that rebalancing also affects higher-order features' effective correlations with labels, and such bias may carry over into models (whether it was originally present or not).  We consider bigrams, as they represent only a slight additional level of complication.

To get a sense of how bigrams overall are affected, we randomly sample 200 bigrams for each of the three successfully rebalanced datasets, selecting uniformly at random among the set of bigrams that appear in at least one instance of each label. We then examine the effect of our (unigram-based) rebalancing of data from table~\ref{tab:rebalancing-real-data} on 
associations in the data between bigram features and labels.  Table~\ref{tab:bigram-balance} shows that in all cases, the average gap between the overall label distribution in the data and the empirical distribution of labels given a bigram \emph{worsens}, despite unigrams' label distributions better reflection of the data's overall label distribution 
(Table~\ref{tab:rebalancing-real-data}) that results from the same reweighted $q$.

This analysis provides a possible explanation for how rebalancing the data with respect to biased unigram features fails to prevent models from learning bias:  the rebalancing didn't correct for biased bigram features, which mislead the model, effectively ``bringing the unigram features'' along with them so that unigram-bias gets learned anyway.  
This is a troubling sign for approaches to bias reduction that focus on data alone, pointing to the need for methods that focus on other aspects of model learning as well.

\section{Methods from Related Work}

Considerable research has posed similar questions of undesirable associations in data manifesting in models, whether through spurious correlations between lexical features and labels  \citep{Tsuchiya2018PerformanceIC, gururangan-etal-2018-annotation, Poliak2018CollectingDN, mccoy-etal-2019-right} or through gender or racial bias \citep{waseem-hovy-2016-hateful, rudinger-etal-2018-gender, stanovsky-etal-2019-evaluating, davidson-etal-2019-racial, sap-etal-2019-risk}.
Out of this large body of work, a few prevailing evaluation methods have emerged.
Foremost among these is assembling a single test set in which a particular bias of interest is lessened and evaluating models' aggregate performance on that test set, such as by excluding instances for which a model that should be too simple to perform the task is correct \citep{gururangan-etal-2018-annotation} or by constructing such a dataset from scratch \citep{mccoy-etal-2019-right}. Similarly, \citet{gardner-etal-2020-evaluating} assemble what is essentially a new, miniature test set (a ``contrast set'') for each human-identified possible category of mistake that a model might make.

We now consider what existing work finds regarding bias in models using these different methods. Overall, we see mixed results. 
\citet{Caliskan2017SemanticsDA} determine that trained word vectors do pick up societal biases from their training corpora. Likewise, \citet{rudinger-etal-2018-gender} find evidence of gender bias in coreference resolution systems, \citet{stanovsky-etal-2019-evaluating} find gender bias in machine translation systems, and \citet{sap-etal-2019-risk} find racial bias in hate speech detection models. However, whether \emph{multiple} attributes' biases in data transfer to models is less clear. For example, \citet{steed-etal-2022-upstream} find that both pretraining data and finetuning data have an effect on biases having to do with gendered pronouns and identity terms that are learned by occupation and toxicity classifiers, but that certain forms of bias reduction in either pretraining or finetuning data don't necessarily overcome bias that the model might pick up from the other.
This is possibly explained by the results of \citet{zhou-srikumar-2022-closer}, who find that data used for finetuning largely distances clusters of textual representations by label without significantly changing other properties of the underlying distribution of data. In a similar vein,
\citet{joshi-he-2022-investigation} find that counterfactually augmented training data can actually exacerbate other spurious correlations in models.

For all the different results reported in this body of literature, there are some typical characteristics of the bias evaluation methodology they apply.  As referenced earlier, it is common for this work to test for a \emph{single} undesirable form of behavior (e.g., biased use of gendered pronouns). For example, \citet{belinkov-etal-2019-dont} focus on whether NLI models ignore input instances' premise, an important problem, but this also simplifies their evaluation, as they doesn't need to consider the potentially disparate impact of their adjusted model on intersecting biases. Another common characteristic is the creation of new and separate test data \citep{mccoy-etal-2019-right, zhang-etal-2019-paws}, on which decreased performance is taken to indicate bias \citep{10.1162/tacl_a_00335, wu-etal-2022-generating}. A concern regarding this strategy, though, is that such test sets very likely still contain (undetected) biases of their own.
Due to the complicated nature of natural language and the highly intertwined features that occur together in text, it is very likely that this will be true regardless of the test set created.

Results using our permutation testing framework indicate the difficulty of removing or mitigating bias from data in a way that corresponds to the mechanisms by which models absorb that bias in practice. This is reminiscent of results from, for example, \citet{gonen-goldberg-2019-lipstick} or \citet{elazar-goldberg-2018-adversarial}, who note that certain ways of seemingly covering up bias still leave traces of that bias in models, and is in line with arguments made by, for example, \citet{eisenstein-2022-informativeness} and \citet{schwartz-stanovsky-2022-limitations}. Further development and testing of hypotheses about how models acquire bias will be important to ensuring that they truly perform the tasks that we intend, and not versions that rely on biased shortcuts in the data.

\section{Conclusion}
We explored how lexical bias in labeled data affects bias in models trained on that data.  Our methodological contribution is a procedure, based on the permutation test, for analyzing biased associations between given features and model predictions, in test data that might itself contain biases.  Our empirical finding is that, in cases where a dataset can be rebalanced to remove most lexical bias, the resulting models remain biased.  This may be related to our observation that the correlations of higher-order (bigram) features with labels actually get \emph{worse} after rebalancing.  We conclude that reducing bias in NLP models may not be achievable by altering existing training data distributions.

\section*{Limitations}

One of the limitations of this work is that we restrict ourselves to examining datasets for supervised learning that contain relatively short instances of text. This likely facilitated the reweighting of data that we wished to perform as an intervention to produce the reweighted data that we study, as the short length of each text effectively capped the number of different lexical features that could cooccur in the same instance. The results we present here might not be representative of lexical feature bias in data with much longer units of text. Also, the fact that the datasets that we used are all in English means that our lexical features were premised on simple whitespace tokenization with punctuation removal; for other languages with a larger variety of reasonable tokenization schemes at varying levels of granularity, the distribution of lexical features, and the resulting conclusions, might look very different.

In addition, apart from the issues we have raised in transferring reduced bias in data to models, we note that an exhaustive list of \textit{all} features that are present in particular data is extremely impractical (and in some cases impossible); any set of features will inevitably leave out some trait of the data, making the reweighting procedure we follow in this work inherently incomprehensive. For those features not included in the problem setup, the measured quality of a returned $q$ distribution will not reflect any changes relevant to those features, although the balance of those features has likely also changed. Even among the features included in the problem input, shifting $q$'s probability mass to improve the balance for one set of features' labels may simultaneously hurt the balance for another.

\section*{Ethics Statement}

This work addresses one piece of the much broader set of questions surrounding how biases---from low-level word associations to high-level social biases---manifest in natural language, and the effects that they have on the models that we train and develop as researchers and practitioners. Parsing out how such biases transfer to models, and when they are harmful, has been and will continue to be key to making progress towards understanding the technologies we create and the scope of what they can or should do.

\section*{Acknowledgments}

The authors appreciate helpful feedback from the anonymous reviewers and members of Noah's ARK at UW and the AllenNLP group at AI2, as well as from Terra Blevins, Yulia Tsvetkov, Lucy Lu Wang, Sheng Wang, and Tim Althoff.

\bibliography{main}
\bibliographystyle{acl_natbib}

\newpage

\appendix

\input{appendices.tex}

\end{document}

%% file: figs/table-balancingdatasets.tex
\begin{table*}[!htbp]
    \centering
    \begin{tabular}{l|c|c|c|c|c}
    
         \multicolumn{1}{c}{} & \multicolumn{1}{c}{$|\pazocal{D}|$} &  \multicolumn{1}{c}{\# Features} & \multicolumn{1}{c}{$|\pazocal{Y}|$} & \multicolumn{1}{c}{\makecell{$\textrm{Err(Uniform)} \quad (\downarrow)$}} & \multicolumn{1}{c}{\makecell{$\textrm{Err(Adjusted }q) \quad (\downarrow)$ }} \\ \hline
         
         \makecell{SNLI} & 549,367 & 3866 & 3 & 0.057 & 0.040 \\ \hline

         \makecell{MNLI} & 392,376 & 6854 & 3 & 0.022 & 0.084   \\  \hline
         
         \makecell{QNLI} & 104,743 & 3770 & 2 & 0.042 & 0.012  \\ \hline
         
         \makecell{QQP} & 363,831 & 4386 & 2 & 0.154 & 0.047  \\ \hline
           
    \end{tabular}
    \caption{The average absolute difference between the empirical fraction of label $y$ in instances with any particular unigram feature $j$ and the total weight given to label $y$ in the full training data, computed over all features and all their label values. Lower is better.}
    \label{tab:rebalancing-real-data}
\end{table*}

%% file: figs/table-pvals.tex
\begin{table*}[!ht]
    \centering
    \begin{tabular}{r|r|r|c}
         \multicolumn{1}{c}{} & \multicolumn{1}{c}{} & \multicolumn{1}{c}{} & \multicolumn{1}{c}{$p$-value(s) for permutation test} \\ \hline
         
         \multirow{6}{*}{Finetuned transformers} & \multirow{2}{*}{\makecell{SNLI}} &  Trained on uniform & $1.9 \times 10^{-35}, \{1.1, 2.2\} \times 10^{-23}$ \\ 
         & & Trained on adjusted $q$ & $\{1.2, 1.7, 3.2\} \times 10^{-14}$ \\ \cline{2-4}
         
         & \multirow{2}{*}{QNLI} & Trained on uniform & $5.7 \times 10^{-3}, \{ 2.4, 5.3\} \times 10^{-2}$ \\
         & & Trained on adjusted $q$ & $\{3.7, 7.6, 2.6\} \times 10^{-1}$ \\  \cline{2-4}
         
         & \multirow{2}{*}{QQP} & Trained on uniform & 
         $2.4 \times 10^{-26}, 2.6 \times 10^{-20}, 2.3 \times 10^{-17}$
         \\
         & & Trained on adjusted $q$ & 
         $7.6 \times 10^{-20}, 5.9 \times 10^{-7}, 1.2 \times 10^{-5}$ 
         \\ \hline

         \multirow{6}{*}{From-scratch LSTM} & \multirow{2}{*}{SNLI} & Trained on uniform & $5.9 \times 10^{-83}$ \\
         & & Trained on adjusted $q$ & $2.0 \times 10^{-75}$ \\ \cline{2-4}

         & \multirow{2}{*}{QNLI} & Trained on uniform & $3.1 \times 10^{-61}$ \\
         & & Trained on adjusted $q$ & $1.6 \times 10^{-10}$ \\ \cline{2-4}

         & \multirow{2}{*}{QQP} & Trained on uniform & Approx. $10^{-638}$ \\
         & & Trained on adjusted $q$ & Approx. $10 ^ {-762}$ \\ \hline
         
    \end{tabular}
    \caption{Exact $p$-values for permutation tests conducted on different models, which check the probability that the usual-gold-label subset of the test data would have at least its observed accuracy if the instances guessed correctly by the model were distributed uniformly at random across the usual and unusual gold-label test subsets. The pretrained model used to initialize each finetuned transformer was RoBERTa-large, and for each pairing of a dataset and a uniform or adjusted weighting of its data in finetuning a transformer, we ran three separate random seeds to observe variance. For each dataset-weighting pairing in training LSTMs from scratch, we used a single random seed.}
    \label{tab:pvals}
\end{table*}

%% file: figs/table-bigrambalance.tex
\begin{table}[!ht]
    \centering
    \begin{tabular}{l|cc}

         \multicolumn{1}{c}{} 
         & \multicolumn{1}{c}{\makecell{$\textrm{Err(Uniform)}  (\downarrow)$}} & \multicolumn{1}{c}{\makecell{$\textrm{Err(Adjusted }q) (\downarrow)$ }} \\ \hline
         
         \makecell{SNLI} 
         & 0.059 & 0.122 \\ \hline
         
         \makecell{QNLI} 
         & 0.134 & 0.173 \\ \hline
         
         \makecell{QQP} 
         & 0.215 & 0.224 \\ \hline

    \end{tabular}
    \caption{
    The average absolute difference between the empirical distribution of label $y$ (in the data) for instances with a \textbf{bigram} feature $j$ and the overall distribution of label $y$ given the full data (we perform this difference elementwise). The calculations over any row in this table are performed over 200 randomly selected bigrams $j$ from that dataset, which are kept consistent across columns. Lower is better.}
    \label{tab:bigram-balance}
\end{table}

%% file: appendices.tex
\section{Appendix}
\label{sec:appendix}

\subsection{List of non-stop-word types most associated with each SNLI label} \label{sec:snli-words}

\subsubsection{Entailment}

These were the 50 word types (after stop words were filtered out) that had the highest $z$-scores for the ``entailment'' label in SNLI:

outside

outdoors

person

near

people

animal

human

humans

least

someone

moving

instrument

something

animals

sport

together

wet

touching

vehicle

things

theres

clothes

multiple

picture

proximity

interacting

physical

using

activity

canine

music

active

musical

object

wears

motion

consuming

clothed

clothing

mammals

working

objects

present

kid

holding

affection

holds

close

instruments

sitted

\subsubsection{Contradiction}

These were the 50 word types (after stop words were filtered out) that had the highest $z$-scores for the ``contradiction'' label in SNLI:

sleeping

nobody

cat

eating

sitting

tv

alone

swimming

asleep

inside

bed

couch

cats

naked

driving

home

empty

eats

car

nothing

running

watching

woman

movie

basketball

nap

television

pool

sleep

anything

moon

beach

man

quietly

laying

room

frowning

sleeps

riding

flying

sits

napping

crying

house

desert

dancing

bench

theater

indoors

pizza

\subsubsection{Neutral}

These were the 50 word types (after stop words were filtered out) that had the highest $z$-scores for the ``neutral'' label in SNLI:

friends

tall

trying

waiting

new

sad

owner

first

competition

going

favorite

friend

winning

vacation

get

date

birthday

wife

work

brothers

ready

party

mother

family

sisters

championship

win

husband

time

fun

siblings

getting

fetch

parents

tired

school

father

best

money

day

married

son

competing

way

wants

professional

trip

likes

show

got

%% file: main.bbl
\begin{thebibliography}{43}
\expandafter\ifx\csname natexlab\endcsname\relax\def\natexlab#1{#1}\fi

\bibitem[{Belinkov et~al.(2019)Belinkov, Poliak, Shieber, Van~Durme, and
  Rush}]{belinkov-etal-2019-dont}
Yonatan Belinkov, Adam Poliak, Stuart Shieber, Benjamin Van~Durme, and
  Alexander Rush. 2019.
\newblock \href {https://doi.org/10.18653/v1/P19-1084} {Don{'}t take the
  premise for granted: Mitigating artifacts in natural language inference}.
\newblock In \emph{Proceedings of the 57th Annual Meeting of the Association
  for Computational Linguistics}, pages 877--891, Florence, Italy. Association
  for Computational Linguistics.

\bibitem[{Bowman et~al.(2015)Bowman, Angeli, Potts, and
  Manning}]{bowman-etal-2015-large}
Samuel~R. Bowman, Gabor Angeli, Christopher Potts, and Christopher~D. Manning.
  2015.
\newblock \href {https://doi.org/10.18653/v1/D15-1075} {A large annotated
  corpus for learning natural language inference}.
\newblock In \emph{Proceedings of the 2015 Conference on Empirical Methods in
  Natural Language Processing}, pages 632--642, Lisbon, Portugal. Association
  for Computational Linguistics.

\bibitem[{Byrd and Lipton(2018)}]{Byrd2018WhatIT}
Jonathon Byrd and Zachary~Chase Lipton. 2018.
\newblock What is the {E}ffect of {I}mportance {W}eighting in {D}eep
  {L}earning?
\newblock In \emph{International Conference on Machine Learning}.

\bibitem[{Caliskan et~al.(2017)Caliskan, Bryson, and
  Narayanan}]{Caliskan2017SemanticsDA}
Aylin Caliskan, Joanna~J. Bryson, and Arvind Narayanan. 2017.
\newblock Semantics derived automatically from language corpora contain
  human-like biases.
\newblock \emph{Science}, 356:183 -- 186.

\bibitem[{Chawla et~al.(2002)Chawla, Bowyer, Hall, and
  Kegelmeyer}]{chawla-smote}
Nitesh~V. Chawla, Kevin~W. Bowyer, Lawrence~O. Hall, and W.~Philip Kegelmeyer.
  2002.
\newblock Smote: Synthetic minority over-sampling technique.
\newblock \emph{J. Artif. Int. Res.}, 16(1):321–357.

\bibitem[{Chen et~al.(2018)Chen, Yang, Liu, Zhao, and
  Jia}]{chen-etal-2018-collective}
Yubo Chen, Hang Yang, Kang Liu, Jun Zhao, and Yantao Jia. 2018.
\newblock \href {https://doi.org/10.18653/v1/D18-1158} {Collective event
  detection via a hierarchical and bias tagging networks with gated multi-level
  attention mechanisms}.
\newblock In \emph{Proceedings of the 2018 Conference on Empirical Methods in
  Natural Language Processing}, pages 1267--1276, Brussels, Belgium.
  Association for Computational Linguistics.

\bibitem[{Clark et~al.(2019)Clark, Yatskar, and
  Zettlemoyer}]{clark-etal-2019-dont}
Christopher Clark, Mark Yatskar, and Luke Zettlemoyer. 2019.
\newblock \href {https://doi.org/10.18653/v1/D19-1418} {Don{'}t {T}ake the
  {E}asy {W}ay {O}ut: {E}nsemble {B}ased {M}ethods for {A}voiding {K}nown
  {D}ataset {B}iases}.
\newblock In \emph{Proceedings of the 2019 Conference on Empirical Methods in
  Natural Language Processing and the 9th International Joint Conference on
  Natural Language Processing (EMNLP-IJCNLP)}, pages 4069--4082, Hong Kong,
  China. Association for Computational Linguistics.

\bibitem[{Davidson et~al.(2019)Davidson, Bhattacharya, and
  Weber}]{davidson-etal-2019-racial}
Thomas Davidson, Debasmita Bhattacharya, and Ingmar Weber. 2019.
\newblock \href {https://doi.org/10.18653/v1/W19-3504} {Racial bias in hate
  speech and abusive language detection datasets}.
\newblock In \emph{Proceedings of the Third Workshop on Abusive Language
  Online}, pages 25--35, Florence, Italy. Association for Computational
  Linguistics.

\bibitem[{Eisenstein(2022)}]{eisenstein-2022-informativeness}
Jacob Eisenstein. 2022.
\newblock \href {https://doi.org/10.18653/v1/2022.naacl-main.321}
  {Informativeness and invariance: Two perspectives on spurious correlations in
  natural language}.
\newblock In \emph{Proceedings of the 2022 Conference of the North American
  Chapter of the Association for Computational Linguistics: Human Language
  Technologies}, pages 4326--4331, Seattle, United States. Association for
  Computational Linguistics.

\bibitem[{Elazar and Goldberg(2018)}]{elazar-goldberg-2018-adversarial}
Yanai Elazar and Yoav Goldberg. 2018.
\newblock \href {https://doi.org/10.18653/v1/D18-1002} {Adversarial removal of
  demographic attributes from text data}.
\newblock In \emph{Proceedings of the 2018 Conference on Empirical Methods in
  Natural Language Processing}, pages 11--21, Brussels, Belgium. Association
  for Computational Linguistics.

\bibitem[{Feldman et~al.(2015)Feldman, Friedler, Moeller, Scheidegger, and
  Venkatasubramanian}]{feldman-certifyingremoving}
Michael Feldman, Sorelle~A. Friedler, John Moeller, Carlos Scheidegger, and
  Suresh Venkatasubramanian. 2015.
\newblock \href {https://doi.org/10.1145/2783258.2783311} {Certifying and
  removing disparate impact}.
\newblock In \emph{Proceedings of the 21th ACM SIGKDD International Conference
  on Knowledge Discovery and Data Mining}, KDD '15, page 259–268, New York,
  NY, USA. Association for Computing Machinery.

\bibitem[{Gardner et~al.(2020)Gardner, Artzi, Basmov, Berant, Bogin, Chen,
  Dasigi, Dua, Elazar, Gottumukkala, Gupta, Hajishirzi, Ilharco, Khashabi, Lin,
  Liu, Liu, Mulcaire, Ning, Singh, Smith, Subramanian, Tsarfaty, Wallace,
  Zhang, and Zhou}]{gardner-etal-2020-evaluating}
Matt Gardner, Yoav Artzi, Victoria Basmov, Jonathan Berant, Ben Bogin, Sihao
  Chen, Pradeep Dasigi, Dheeru Dua, Yanai Elazar, Ananth Gottumukkala, Nitish
  Gupta, Hannaneh Hajishirzi, Gabriel Ilharco, Daniel Khashabi, Kevin Lin,
  Jiangming Liu, Nelson~F. Liu, Phoebe Mulcaire, Qiang Ning, Sameer Singh,
  Noah~A. Smith, Sanjay Subramanian, Reut Tsarfaty, Eric Wallace, Ally Zhang,
  and Ben Zhou. 2020.
\newblock \href {https://doi.org/10.18653/v1/2020.findings-emnlp.117}
  {Evaluating models{'} local decision boundaries via contrast sets}.
\newblock In \emph{Findings of the Association for Computational Linguistics:
  EMNLP 2020}, pages 1307--1323, Online. Association for Computational
  Linguistics.

\bibitem[{Gardner et~al.(2021)Gardner, Merrill, Dodge, Peters, Ross, Singh, and
  Smith}]{gardner-etal-2021-competency}
Matt Gardner, William Merrill, Jesse Dodge, Matthew Peters, Alexis Ross, Sameer
  Singh, and Noah~A. Smith. 2021.
\newblock \href {https://doi.org/10.18653/v1/2021.emnlp-main.135} {Competency
  problems: On finding and removing artifacts in language data}.
\newblock In \emph{Proceedings of the 2021 Conference on Empirical Methods in
  Natural Language Processing}, pages 1801--1813, Online and Punta Cana,
  Dominican Republic. Association for Computational Linguistics.

\bibitem[{Gonen and Goldberg(2019)}]{gonen-goldberg-2019-lipstick}
Hila Gonen and Yoav Goldberg. 2019.
\newblock \href {https://doi.org/10.18653/v1/N19-1061} {Lipstick on a pig:
  {D}ebiasing methods cover up systematic gender biases in word embeddings but
  do not remove them}.
\newblock In \emph{Proceedings of the 2019 Conference of the North {A}merican
  Chapter of the Association for Computational Linguistics: Human Language
  Technologies, Volume 1 (Long and Short Papers)}, pages 609--614, Minneapolis,
  Minnesota. Association for Computational Linguistics.

\bibitem[{Gowda et~al.(2021)Gowda, Joshi, Zhang, and
  Ghassemi}]{Gowda2021PullingUB}
Sindhu C.~M. Gowda, Shalmali Joshi, Haoran Zhang, and Marzyeh Ghassemi. 2021.
\newblock Pulling up by the causal bootstraps: Causal data augmentation for
  pre-training debiasing.
\newblock \emph{Proceedings of the 30th ACM International Conference on
  Information \& Knowledge Management}.

\bibitem[{Gururangan et~al.(2018)Gururangan, Swayamdipta, Levy, Schwartz,
  Bowman, and Smith}]{gururangan-etal-2018-annotation}
Suchin Gururangan, Swabha Swayamdipta, Omer Levy, Roy Schwartz, Samuel Bowman,
  and Noah~A. Smith. 2018.
\newblock \href {https://doi.org/10.18653/v1/N18-2017} {Annotation artifacts in
  natural language inference data}.
\newblock In \emph{Proceedings of the 2018 Conference of the North {A}merican
  Chapter of the Association for Computational Linguistics: Human Language
  Technologies, Volume 2 (Short Papers)}, pages 107--112, New Orleans,
  Louisiana. Association for Computational Linguistics.

\bibitem[{Joshi and He(2022)}]{joshi-he-2022-investigation}
Nitish Joshi and He~He. 2022.
\newblock \href {https://doi.org/10.18653/v1/2022.acl-long.256} {An
  investigation of the (in)effectiveness of counterfactually augmented data}.
\newblock In \emph{Proceedings of the 60th Annual Meeting of the Association
  for Computational Linguistics (Volume 1: Long Papers)}, pages 3668--3681,
  Dublin, Ireland. Association for Computational Linguistics.

\bibitem[{Kamiran and Calders(2012)}]{kamiran2012data}
Faisal Kamiran and Toon Calders. 2012.
\newblock Data preprocessing techniques for classification without
  discrimination.
\newblock \emph{Knowledge and information systems}, 33(1):1--33.

\bibitem[{Karimi~Mahabadi et~al.(2020)Karimi~Mahabadi, Belinkov, and
  Henderson}]{karimi-mahabadi-etal-2020-end}
Rabeeh Karimi~Mahabadi, Yonatan Belinkov, and James Henderson. 2020.
\newblock \href {https://doi.org/10.18653/v1/2020.acl-main.769} {{E}nd-to-{E}nd
  {B}ias {M}itigation by {M}odelling {B}iases in {C}orpora}.
\newblock In \emph{Proceedings of the 58th Annual Meeting of the Association
  for Computational Linguistics}, pages 8706--8716, Online. Association for
  Computational Linguistics.

\bibitem[{Lee et~al.(2021)Lee, Won, Kim, Lee, Park, and
  Jung}]{Lee2021CrossAugAC}
Minwoo Lee, Seungpil Won, Juae Kim, Hwanhee Lee, Cheoneum Park, and Kyomin
  Jung. 2021.
\newblock Crossaug: A contrastive data augmentation method for debiasing fact
  verification models.
\newblock \emph{Proceedings of the 30th ACM International Conference on
  Information \& Knowledge Management}.

\bibitem[{Liu et~al.(2019)Liu, Ott, Goyal, Du, Joshi, Chen, Levy, Lewis,
  Zettlemoyer, and Stoyanov}]{roberta}
Yinhan Liu, Myle Ott, Naman Goyal, Jingfei Du, Mandar Joshi, Danqi Chen, Omer
  Levy, Mike Lewis, Luke Zettlemoyer, and Veselin Stoyanov. 2019.
\newblock \href {https://doi.org/10.48550/ARXIV.1907.11692} {Ro{BERT}a: {A
  R}obustly {O}ptimized {BERT P}retraining {A}pproach}.

\bibitem[{McCoy et~al.(2019)McCoy, Pavlick, and Linzen}]{mccoy-etal-2019-right}
Tom McCoy, Ellie Pavlick, and Tal Linzen. 2019.
\newblock \href {https://doi.org/10.18653/v1/P19-1334} {Right for the wrong
  reasons: Diagnosing syntactic heuristics in natural language inference}.
\newblock In \emph{Proceedings of the 57th Annual Meeting of the Association
  for Computational Linguistics}, pages 3428--3448, Florence, Italy.
  Association for Computational Linguistics.

\bibitem[{Mehrabi et~al.(2021)Mehrabi, Morstatter, Saxena, Lerman, and
  Galstyan}]{mehrabi2021survey}
Ninareh Mehrabi, Fred Morstatter, Nripsuta Saxena, Kristina Lerman, and Aram
  Galstyan. 2021.
\newblock A survey on bias and fairness in machine learning.
\newblock \emph{ACM Computing Surveys (CSUR)}, 54(6):1--35.

\bibitem[{Mikolov et~al.(2013)Mikolov, Chen, Corrado, and
  Dean}]{mikolovword2vec}
Tomas Mikolov, Kai Chen, Greg Corrado, and Jeffrey Dean. 2013.
\newblock \href {https://doi.org/10.48550/ARXIV.1301.3781} {Efficient
  {E}stimation of {W}ord {R}epresentations in {V}ector {S}pace}.

\bibitem[{Phipson and Smyth(2010)}]{PhipsonSmyth+2010}
Belinda Phipson and Gordon~K Smyth. 2010.
\newblock \href {https://doi.org/doi:10.2202/1544-6115.1585} {Permutation
  p-values should never be zero: Calculating exact p-values when permutations
  are randomly drawn}.
\newblock \emph{Statistical Applications in Genetics and Molecular Biology},
  9(1).

\bibitem[{Poliak et~al.(2018)Poliak, Haldar, Rudinger, Hu, Pavlick, White, and
  Durme}]{Poliak2018CollectingDN}
Adam Poliak, Aparajita Haldar, Rachel Rudinger, J.~Edward Hu, Ellie Pavlick,
  Aaron~Steven White, and Benjamin~Van Durme. 2018.
\newblock Collecting diverse natural language inference problems for sentence
  representation evaluation.
\newblock In \emph{Conference on Empirical Methods in Natural Language
  Processing}.

\bibitem[{Rudinger et~al.(2018)Rudinger, Naradowsky, Leonard, and
  Van~Durme}]{rudinger-etal-2018-gender}
Rachel Rudinger, Jason Naradowsky, Brian Leonard, and Benjamin Van~Durme. 2018.
\newblock \href {https://doi.org/10.18653/v1/N18-2002} {Gender bias in
  coreference resolution}.
\newblock In \emph{Proceedings of the 2018 Conference of the North {A}merican
  Chapter of the Association for Computational Linguistics: Human Language
  Technologies, Volume 2 (Short Papers)}, pages 8--14, New Orleans, Louisiana.
  Association for Computational Linguistics.

\bibitem[{Sap et~al.(2019)Sap, Card, Gabriel, Choi, and
  Smith}]{sap-etal-2019-risk}
Maarten Sap, Dallas Card, Saadia Gabriel, Yejin Choi, and Noah~A. Smith. 2019.
\newblock \href {https://doi.org/10.18653/v1/P19-1163} {The risk of racial bias
  in hate speech detection}.
\newblock In \emph{Proceedings of the 57th Annual Meeting of the Association
  for Computational Linguistics}, pages 1668--1678, Florence, Italy.
  Association for Computational Linguistics.

\bibitem[{Schwartz and Stanovsky(2022)}]{schwartz-stanovsky-2022-limitations}
Roy Schwartz and Gabriel Stanovsky. 2022.
\newblock \href {https://doi.org/10.18653/v1/2022.findings-naacl.168} {On the
  limitations of dataset balancing: The lost battle against spurious
  correlations}.
\newblock In \emph{Findings of the Association for Computational Linguistics:
  NAACL 2022}, pages 2182--2194, Seattle, United States. Association for
  Computational Linguistics.

\bibitem[{Stanovsky et~al.(2019)Stanovsky, Smith, and
  Zettlemoyer}]{stanovsky-etal-2019-evaluating}
Gabriel Stanovsky, Noah~A. Smith, and Luke Zettlemoyer. 2019.
\newblock \href {https://doi.org/10.18653/v1/P19-1164} {Evaluating gender bias
  in machine translation}.
\newblock In \emph{Proceedings of the 57th Annual Meeting of the Association
  for Computational Linguistics}, pages 1679--1684, Florence, Italy.
  Association for Computational Linguistics.

\bibitem[{Steed et~al.(2022)Steed, Panda, Kobren, and
  Wick}]{steed-etal-2022-upstream}
Ryan Steed, Swetasudha Panda, Ari Kobren, and Michael Wick. 2022.
\newblock \href {https://doi.org/10.18653/v1/2022.acl-long.247} {{U}pstream
  {M}itigation {I}s \textit{ {N}ot} {A}ll {Y}ou {N}eed: {T}esting the {B}ias
  {T}ransfer {H}ypothesis in {P}re-{T}rained {L}anguage {M}odels}.
\newblock In \emph{Proceedings of the 60th Annual Meeting of the Association
  for Computational Linguistics (Volume 1: Long Papers)}, pages 3524--3542,
  Dublin, Ireland. Association for Computational Linguistics.

\bibitem[{Tsuchiya(2018)}]{Tsuchiya2018PerformanceIC}
Masatoshi Tsuchiya. 2018.
\newblock Performance impact caused by hidden bias of training data for
  recognizing textual entailment.
\newblock \emph{ArXiv}, abs/1804.08117.

\bibitem[{Tu et~al.(2020)Tu, Lalwani, Gella, and He}]{10.1162/tacl_a_00335}
Lifu Tu, Garima Lalwani, Spandana Gella, and He~He. 2020.
\newblock \href {https://doi.org/10.1162/tacl_a_00335} {{An Empirical Study on
  Robustness to Spurious Correlations using Pre-trained Language Models}}.
\newblock \emph{Transactions of the Association for Computational Linguistics},
  8:621--633.

\bibitem[{Wang et~al.(2018)Wang, Singh, Michael, Hill, Levy, and
  Bowman}]{wang-etal-2018-glue}
Alex Wang, Amanpreet Singh, Julian Michael, Felix Hill, Omer Levy, and Samuel
  Bowman. 2018.
\newblock \href {https://doi.org/10.18653/v1/W18-5446} {{GLUE}: A multi-task
  benchmark and analysis platform for natural language understanding}.
\newblock In \emph{Proceedings of the 2018 {EMNLP} Workshop {B}lackbox{NLP}:
  Analyzing and Interpreting Neural Networks for {NLP}}, pages 353--355,
  Brussels, Belgium. Association for Computational Linguistics.

\bibitem[{Waseem and Hovy(2016)}]{waseem-hovy-2016-hateful}
Zeerak Waseem and Dirk Hovy. 2016.
\newblock \href {https://doi.org/10.18653/v1/N16-2013} {Hateful symbols or
  hateful people? predictive features for hate speech detection on {T}witter}.
\newblock In \emph{Proceedings of the {NAACL} Student Research Workshop}, pages
  88--93, San Diego, California. Association for Computational Linguistics.

\bibitem[{Williams et~al.(2018)Williams, Nangia, and
  Bowman}]{williams-etal-2018-broad}
Adina Williams, Nikita Nangia, and Samuel Bowman. 2018.
\newblock \href {https://doi.org/10.18653/v1/N18-1101} {A broad-coverage
  challenge corpus for sentence understanding through inference}.
\newblock In \emph{Proceedings of the 2018 Conference of the North {A}merican
  Chapter of the Association for Computational Linguistics: Human Language
  Technologies, Volume 1 (Long Papers)}, pages 1112--1122, New Orleans,
  Louisiana. Association for Computational Linguistics.

\bibitem[{Wu et~al.(2022)Wu, Gardner, Stenetorp, and
  Dasigi}]{wu-etal-2022-generating}
Yuxiang Wu, Matt Gardner, Pontus Stenetorp, and Pradeep Dasigi. 2022.
\newblock \href {https://doi.org/10.18653/v1/2022.acl-long.190} {Generating
  data to mitigate spurious correlations in natural language inference
  datasets}.
\newblock In \emph{Proceedings of the 60th Annual Meeting of the Association
  for Computational Linguistics (Volume 1: Long Papers)}, pages 2660--2676,
  Dublin, Ireland. Association for Computational Linguistics.

\bibitem[{Zellers et~al.(2019)Zellers, Holtzman, Rashkin, Bisk, Farhadi,
  Roesner, and Choi}]{zellers-grover}
Rowan Zellers, Ari Holtzman, Hannah Rashkin, Yonatan Bisk, Ali Farhadi,
  Franziska Roesner, and Yejin Choi. 2019.
\newblock \href
  {https://proceedings.neurips.cc/paper/2019/file/3e9f0fc9b2f89e043bc6233994dfcf76-Paper.pdf}
  {Defending against neural fake news}.
\newblock In \emph{Advances in Neural Information Processing Systems},
  volume~32. Curran Associates, Inc.

\bibitem[{Zhai et~al.(2023)Zhai, Dan, Kolter, and
  Ravikumar}]{Zhai2022UnderstandingWG}
Runtian Zhai, Chen Dan, J.~Zico Kolter, and Pradeep Ravikumar. 2023.
\newblock Understanding {W}hy {G}eneralized {R}eweighting {D}oes {N}ot
  {I}mprove {O}ver {ERM}.
\newblock In \emph{Proceedings of the International Conference on Learning
  Representations}.

\bibitem[{Zhang and Sang(2020)}]{Zhang2020TowardsAP}
Yi~Zhang and Jitao Sang. 2020.
\newblock Towards accuracy-fairness paradox: Adversarial example-based data
  augmentation for visual debiasing.
\newblock \emph{Proceedings of the 28th ACM International Conference on
  Multimedia}.

\bibitem[{Zhang et~al.(2019)Zhang, Baldridge, and He}]{zhang-etal-2019-paws}
Yuan Zhang, Jason Baldridge, and Luheng He. 2019.
\newblock \href {https://doi.org/10.18653/v1/N19-1131} {{PAWS}: Paraphrase
  adversaries from word scrambling}.
\newblock In \emph{Proceedings of the 2019 Conference of the North {A}merican
  Chapter of the Association for Computational Linguistics: Human Language
  Technologies, Volume 1 (Long and Short Papers)}, pages 1298--1308,
  Minneapolis, Minnesota. Association for Computational Linguistics.

\bibitem[{Zhao et~al.(2018)Zhao, Wang, Yatskar, Ordonez, and
  Chang}]{zhao-etal-2018-gender}
Jieyu Zhao, Tianlu Wang, Mark Yatskar, Vicente Ordonez, and Kai-Wei Chang.
  2018.
\newblock \href {https://doi.org/10.18653/v1/N18-2003} {Gender bias in
  coreference resolution: Evaluation and debiasing methods}.
\newblock In \emph{Proceedings of the 2018 Conference of the North {A}merican
  Chapter of the Association for Computational Linguistics: Human Language
  Technologies, Volume 2 (Short Papers)}, pages 15--20, New Orleans, Louisiana.
  Association for Computational Linguistics.

\bibitem[{Zhou and Srikumar(2022)}]{zhou-srikumar-2022-closer}
Yichu Zhou and Vivek Srikumar. 2022.
\newblock \href {https://doi.org/10.18653/v1/2022.acl-long.75} {A closer look
  at how fine-tuning changes {BERT}}.
\newblock In \emph{Proceedings of the 60th Annual Meeting of the Association
  for Computational Linguistics (Volume 1: Long Papers)}, pages 1046--1061,
  Dublin, Ireland. Association for Computational Linguistics.

\end{thebibliography}
